\documentclass{article}

\usepackage[utf8]{inputenc} 
\usepackage[T1]{fontenc}    
\usepackage{hyperref}       
\usepackage{url}            
\usepackage{booktabs}       
\usepackage{amsfonts}       
\usepackage{nicefrac}       
\usepackage{microtype}      
\usepackage{adjustbox}
\usepackage{todonotes}

\usepackage[utf8]{inputenc} 
\usepackage[T1]{fontenc}    
\usepackage{hyperref}       
\usepackage{url}            
\usepackage{booktabs}       
\usepackage{amsfonts}       
\usepackage{nicefrac}       
\usepackage{microtype}      
\usepackage{graphicx}
\usepackage{amsthm}
\usepackage{amsmath}
\usepackage[english]{babel}
\usepackage[T1]{fontenc}
\usepackage[utf8]{inputenc}
\usepackage{authblk}
\usepackage{bbm}
\usepackage{wrapfig}
\usepackage[nonatbib, final]{corl_2021}

\author[1]{Hugo Caselles-Dupr\'e}

\author[2]{Michael Garcia-Ortiz}

\author[1]{David Filliat}

\affil[1]{U2IS, ENSTA Paris, Institut Polytechnique de Paris \& INRIA Flowers}
\affil[2]{CitAI, SMCSE, City University of London}

\author{Hugo Caselles-Dupr\'e}

\title{Are standard Object Segmentation models sufficient for Learning Affordance Segmentation?}

\begin{document}

\maketitle

\begin{abstract}

Affordances are the possibilities of actions the environment offers to the individual. Ordinary objects (hammer, knife) usually have many affordances (grasping, pounding, cutting), and detecting these allow artificial agents to understand what are their possibilities in the environment, with obvious application in Robotics. Proposed benchmarks and state-of-the-art prediction models for supervised affordance segmentation are usually modifications of popular object segmentation models such as Mask R-CNN. We observe that theoretically, these popular object segmentation methods should be sufficient for detecting affordances masks. So we ask the question: is it necessary to tailor new architectures to the problem of learning affordances? We show that applying the out-of-the-box Mask R-CNN to the problem of affordances segmentation outperforms the current state-of-the-art. We conclude that the problem of supervised affordance segmentation is included in the problem of object segmentation and argue that better benchmarks for affordance learning should include action capacities.

\end{abstract}

\section{Introduction}

Scene understanding involves numerous tasks, including recognizing which objects are present, localizing the objects in 2D and 3D, determining the objects’ and scene’s attributes, characterizing relationships between objects and providing a semantic description of the scene. Understanding a visual scene is one of the primary goals of computer vision. In this work, we compare two important tasks that allow scene understanding: object segmentation and affordance segmentation. 

\begin{figure}[!ht]
  \centering
{\includegraphics[scale=0.8]{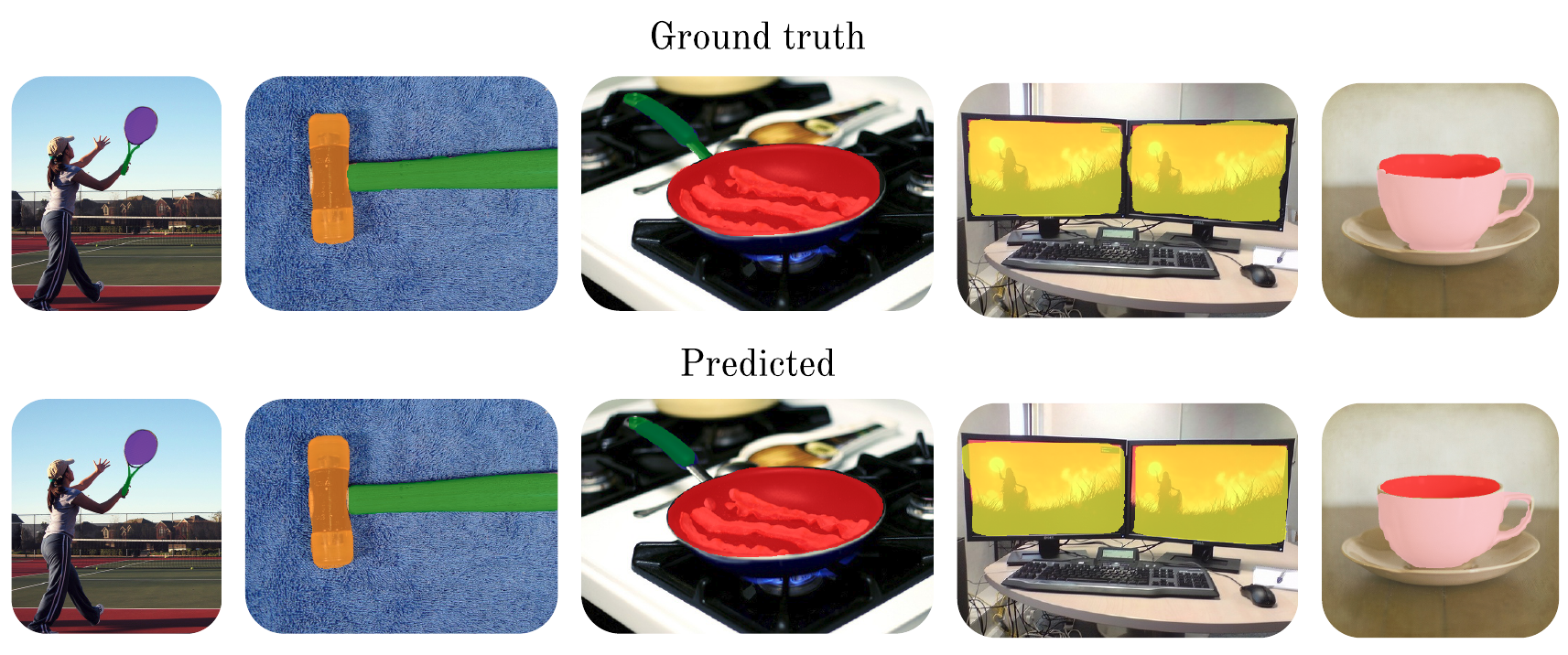}} \hfill
   \caption{Illustration of ground-truth and predicted affordance segmentation masks using Mask R-CNN for the IIT-AFF dataset \cite{nguyen2017object}. \label{fig:res_illu}}
\end{figure}

On the one hand, object segmentation aims at predicting masks for objects that appear in a visual scene, and uses algorithms most of the time trained with supervision from  a labeled dataset of images with the corresponding objects segmentation. The algorithms rely on largely annotated databases of object classes that allow to train fast and accurate predictive models for bounding boxes and segmentation masks. Pascal VOC \cite{everingham2015pascal} and COCO \cite{lin2014microsoft} are examples of the most used datasets, and popular methods like Mask R-CNN \cite{he2017mask} or YOLO \cite{redmon2016you} and their variants have shown to be very efficient at solving object detection and segmentation tasks, providing numerous real-life applications such as face detection, people counting, vehicle detection, or product recognition for manufacturing industries.

On the other hand, affordance segmentation in computer vision aims at characterizing the functionality of an object. Affordances is a term created by Gibson \cite{gibson2014ecological} that describes the opportunity for action that elements in the environment provide to an agent. Depending on the situation, an object can have different affordances, for instance a chair can be used as a seat or as something to stand on to reach for objects that are high in altitude. Affordances require a relationship in which the environment and the agent can work together. Recent work have focused on creating labeled dataset for affordance learning \cite{nguyen2017object, myers2015affordance}, and with that came several deep learning algorithms that provide methods for the automatic inference of affordances \cite{nguyen2016detecting, koppula2013learning, do2018affordancenet, chuang2018learning, fang2018demo2vec, zeng2018robotic}. 

Object segmentation and affordance segmentation are obviously related, being two visual segmentation tasks. In our work, we take a closer look at the relationship between the two tasks, and ask: does supervised object segmentation methods suffice for affordance segmentation? Indeed, object segmentation is a more general problem, and has been studied widely in the last decade, taking advantage of the computer vision revolution brought by deep learning methods. The result is general visual segmentation methods that are easily re-usable, fast and efficient. 

There are existing work that aim at adapting and tailoring those methods for affordance segmentation. We challenge this approach by experimenting with out-the-box pretrained methods, namely Mask R-CNN, and fine-tuning them to two affordance segmentation datasets. We compare our results with the previous state-of-the-art in affordance segmentation, including methods that are specifically modified for this task, and we surprisingly surpass those methods by a large margin, setting a new state-of-the-art on both datasets. From this unexpected result, we deduce that, in the current research landscape, the problem of learning affordance segmentation is included in the problem of learning object segmentation. Finally, from the conclusion obtained with our experiments, we provide a perspective on the importance of affordance prediction and how affordance segmentation benchmarks could be improved to better represent the task and its complexity.

Our contributions are the following:

\begin{itemize}
    \item By fine-tuning a pre-trained Mask R-CNN segmentation algorithm, we set a new state-of-the-art performance on two affordance segmentation datasets, surpassing previous approaches by a large margin.
    \item We use our performance comparison to deduce that the problem of supervised affordance segmentation is a sub-problem of supervised object segmentation.
    \item We provide guidelines on how affordance benchmarks should be built to foster progress in affordance detection.
\end{itemize}

\section{Supervised Object Segmentation vs. Supervised Affordance Segmentation}

In this section we introduce and compare the two tasks of interest in our study: standard supervised object segmentation (OS) and supervised affordance segmentation (AS).

\subsection{Supervised Object Segmentation}

\quad \textbf{Task description.} In OS, the goal is to predict a segmentation mask of the objects present in the scene. Those objects might be mugs, tables or chairs, but also persons or animals, or parts of them such as eyes or mug handles. The algorithm should produce a list of object categories present in the image along with a labelling of the pixels corresponding to each object. From these segmentation masks, an axis-aligned bounding box indicating the position and scale of every instance of each object category can be extracted. In order to compare the predicted masks and the ground-truth masks, there exists several metrics usually based on a trade-off between precision (fraction of relevant instances among the retrieved instances) and recall (fraction of relevant instances that were retrieved).

\quad \textbf{State-of-the-art models.} The R-CNN model family \cite{girshick2014rich,girshick2015fast,jiang2017face,he2017mask} is one of the most praised approach for object detection tasks. The first iteration was R-CNN \cite{girshick2014rich}. The idea was to generate and extract category independent region proposals, e.g. candidate bounding boxes, using a region-proposal algorithm such as selective search \cite{uijlings2013selective} and then extract feature from each candidate region before finally classifying features as one of the known class. R-CNN is one of the first successful application of convolutional neural networks to the problem of object localization, detection, and segmentation, however it was a slow and multi-stage pipeline.

To overcome this, Fast R-CNN \cite{girshick2015fast} proposed as a single model instead of a pipeline to learn and output regions and classifications directly. It takes the image and a set of region proposals as input that are passed through a deep convolutional neural network. A pre-trained CNN, such as a VGG-16, is used for feature extraction. Then comes a Region of Interest Pooling Layer, or RoI Pooling, that extracts features specific for a given input candidate region. Then the model bifurcates into two outputs, one for the class prediction via a softmax layer, and another with a linear output for the bounding box. This process is then repeated multiple times for each region of interest in a given image.

Fast R-CNN still requires a region proposal algorithm, and thus the next iteration Faster R-CNN \cite{jiang2017face} introduces a novel architecture designed to both propose and refine region proposals as part of the training process, referred to as a Region Proposal Network, or RPN. These regions are then used in concert with a Fast R-CNN model in a single model design. These improvements both reduce the number of region proposals and accelerate the test-time operation of the model to near real-time with then state-of-the-art performance.

Finally, Mask R-CNN \cite{he2017mask} completes Faster R-CNN with binary mask prediction. Mask R-CNN is simply composed of Faster R-CNN to which a third branch that outputs the object mask — which is a binary mask that indicates the pixels where the object is in the bounding box - is added. The additional mask output is distinct from the class and box outputs, requiring extraction of much finer spatial layout of an object. To do this Mask RCNN uses the Fully Convolution Network (FCN) \cite{long2015fully}. Mask R-CNN is a staple in object localization, detection and segmentation, widely used in practice. Novel approach have succeded in surpassing the performance of Mask R-CNN \cite{bolya2019yolact}, however the model is still influential as it is used a strong baseline or building block in many computer vision fields, including affordance segmentation which we will review next.

\begin{figure}
    \centering
    \includegraphics[scale=0.4]{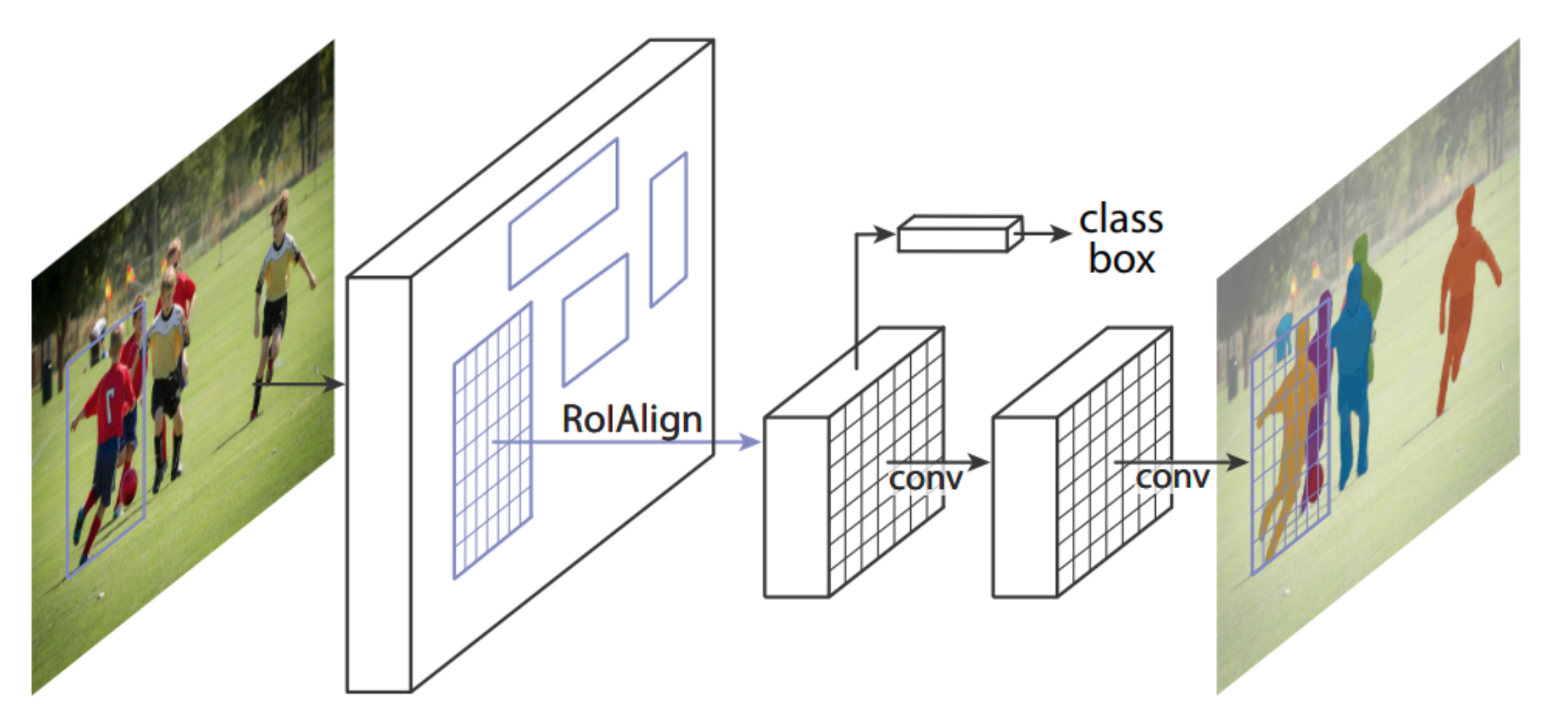}
    \caption{The Mask R-CNN framework for object segmentation. Figure source: Mask R-CNN original paper \cite{he2017mask}.}
    \label{fig:mask-rcnn}
\end{figure}

\subsection{Supervised Affordance Segmentation}

\quad \textbf{Task description.} In AS, the goal is to 1) detect objects and 2) their affordances. The first part is exactly what is described above in OS. The second part implies predicting pixel-wise segmentation masks that represents the utility, or affordance, of the object. In one object, there are usually several affordances to predict. For instance, the blade of the knife represents the cutting affordance and the handle of the cup represents the grasping affordance. From these affordances masks can be extracted a bounding box, just like with object masks. The metrics used to evaluate prediction are identical to the one used in OS. Note that learning affordances in a supervised manner is not the only possible approach, we discuss other approaches in Sec.\ref{sec:discu}.

\quad \textbf{State-of-the-art models.}

There are existing methods based on Mask R-CNN that adapt the architecture to the problem of learning affordance segmentation. State-of-the-art methods are based on an architecture called AffordanceNet \cite{do2018affordancenet}. This architecture builds on Mask R-CNN by adding three modifications: a sequence of deconvolutional layers, a robust resizing strategy, and a new loss function. First, they add a sequence of deconvolution layers to upsample the RoI feature map (resulting from the RPN) to high resolution in order to obtain a smooth and fine affordance map. Second, they implement a resizing strategy based on multiple thresholds. Using a single threshold value, which is done in Mask R-CNN, does not work in their affordance detection problem since they have multiple affordance classes in each object bounding box. Finally, they adapt the loss function of Mask R-CNN to support multiple affordance labels, compared to the standard case with Mask R-CNN where the pixels are either background or foreground in the predicted bounding box of an object. 

In a follow-up study of AffordanceNet performances \cite{minhlearning}, the authors propose another modification for increased performance: multiple alignment. Alignment in Mask R-CNN is usually only applied after the RPN feature extraction stage. Multiple alignment aggregates features from deep stages to the affordance mask branch. They argue that using a single alignment process can lead to inefficiency in segmenting affordances.

\subsection{Overlap between the two tasks}

The two tasks of supervised object segmentation and supervised affordance segmentation are similar because they entail predicting segmentation masks at a pixel level, using the same training paradigm: computation of a loss with respect to the ground-truth labels, and backpropagating the error through the network in order to minimize the loss. 

One motivation of research work on affordance segmentation \cite{do2018affordancenet, minhlearning} is to jointly learn the object label and the affordances within the detected object. Two reasons for this design are 1) improved computational efficiency (especially if we want the model to be able to run on real robots) and 2) a more natural way of detecting affordances (finding the object, and then recognizing its affordances). On the contrary, we propose to directly predict the affordances, without intermediate object detection, and treat object detection as a separated problem if needed.

We argue that even if theoretically the motivation of jointly detecting objects and affordances makes sense, practically it might not be necessary to frame the problem this way. First, the potential issue that decoupling object detection and affordance detection is computationally too slow does not exist anymore because of the recent development in object detection. Studies have shown that Mask R-CNN can run at 10 frame per second (FPS), and many recent object detection methods can run faster than 40 FPS \cite{bolya2019yolact}. This means that having an object detection model and an object affordance model run jointly at high FPS is possible. Essentially, this problem might have existed before 2016, but it is not an issue anymore. 

Secondly, it is not clear why having object segmentation be followed with affordance segmentation would necessary help in the actual performance of the model. While this idea might be appealing in theory, there are no real reason, in practice, not to decouple the task in two parts if this results in a better performing model. Especially since papers that proposes new architectures for affordance segmentation report their results in terms of affordance mask prediction accuracy: if a model A outperforms a model B on this criterion, it is sufficient to call it superior. Hence, any method capable of surpassing the state-of-the-art in affordance mask prediction accuracy is considered superior.

\quad \textbf{Research question.} This analysis naturally raises the question of the need for tailoring new architectures to the problem of affordance segmentation. Can strong object segmentation baselines be applied to the affordance segmentation task? In the remainder of the paper, we experiment with Mask R-CNN on two affordance segmentation benchmarks and show that these baselines can largely surpass approaches tailored for affordance segmentation, hence proving our point: the supervised affordance segmentation task in included in the supervised object segmentation task.

\section{Experimental setup}

We describe the experimental setup used in our paper: datasets, metric, methods and baselines.

\subsection{Datasets and metric}

\quad \textbf{Datasets.} The IIT-AFF dataset \cite{nguyen2017object} consists of $8835$ real-world images. This dataset is composed of around 60\% of the images from ImageNet dataset \cite{deng2009imagenet}, while the rest images are taken by the authors from cluttered scenes. In particular, this dataset contains 10 object categories, 9 affordance classes, $14642$ object bounding boxes, and $24677$ affordance regions at pixel level. We use the standard split as in \cite{nguyen2017object} to train our network (i.e. 70\% for training and 30\% for testing).

The UMD dataset \cite{myers2015affordance} contains around $30k$ RGB-D images of daily kitchen, workshop, and garden objects. The RGB-D images of this dataset were captured from a Kinect camera on a rotating table in a clutter-free setup. This dataset has 7 affordance classes and 17 object categories. We use only the RGB images of this dataset and follow the split in \cite{myers2015affordance} to train and test our network.

\quad \textbf{Metric.} As the standard practice, we use the $F^{w}_{\beta}$ metric \cite{margolin2014evaluate} to evaluate the affordance detection results. This is standard practice in the field of affordance segmentation \cite{do2018affordancenet, minhlearning, nguyen2017object, nguyen2016detecting}. This metric estimates the pixel-to-pixel similarity between a predicted foreground and a ground-truth affordance mask. It is computed as following: 

$$ F^{w}_{\beta} = (1+\beta^2) \cdot \frac{Precision_w \times Recall_w}{\beta^2 \cdot Precision_w + Recall_w}$$

where $Precision_w$ and $Recall_w$ is the weighted Precision and weighted Recall as described in \cite{margolin2014evaluate}. In our experiments, we employ standard values for the parameters of this metric: $\beta=1$, $\sigma^2=5$ and $\alpha=\frac{ln(0.5)}{5}$, see \cite{margolin2014evaluate} for details on these parameters. We not only estimate the average performance across all affordance classes, but we also evaluate performance within each affordance class (see Tables \ref{tb_result_iit} and \ref{tb_result_umd}). This metric estimates the pixel-to-pixel similarity between a predicted foreground and a ground-truth affordance mask.

\subsection{Mask R-CNN implementation}

We use the implementation of Mask R-CNN provided in the well-established MMDetection library \cite{mmdetection}. We select the ResNet-50 backbone, and use a pre-trained model on the COCO dataset \cite{lin2014microsoft}. All hyperparameters and pre-processing of images and segmentation masks used in the training of this model are available in the configuration file provided in the library\footnote{\href{https://github.com/open-mmlab/mmdetection/blob/master/configs/mask_rcnn/mask_rcnn_r50_caffe_fpn_1x_coco.py}{Link to the configuration file of the pre-trained Mask R-CNN.}}. 

We fine-tune this pre-trained Mask R-CNN on the IIT-AFF and UMD datasets. IIT-AFF has 9 affordance labels (\texttt{contain}, \texttt{cut}, \texttt{display}, \texttt{engine}, \texttt{grasp}, \texttt{hit}, \texttt{pound}, \texttt{support} and \texttt{w-grasp}), while the UMD dataset has 7 (\texttt{grasp}, \texttt, \texttt{cut}, \texttt{contain}, \texttt{support}, \texttt{scoop} and \texttt{pound}). Hence, we set the number of trainable prediction heads to the number of affordance classes in the fine-tuning procedure, which is standard practice with Mask R-CNN fine-tuning. The pre-processing of the images and the segmentation masks is identical to the pre-processing used during training. We set the number of epochs to $50$ for the IIT-AFF and $12$ for the UMD dataset. For fine-tuning, we use a Stochastic Gradient Descent (SGD) \cite{robbins1951stochastic} optimizer with a learning of $0.02$ (usually smaller than used during pre-training on COCO dataset \cite{lin2014microsoft}), with a linear warmup from $0.001$ during the first 500 iterations.

For the affordance segmentation mask inference, Mask R-CNN outputs a number of potential masks for each affordance label, with a confidence probability. We compare the most probable mask of each affordance label where there is at least one prediction, with the ground-truth mask, using the aforementioned $F^{w}_{\beta}$ metric. This is equivalent to the inference procedures used in other affordance segmentation papers \cite{do2018affordancenet, minhlearning}.

\subsection{Baselines}

We compare our approach to numerous baselines, including the current state-of-the-art models. This includes: DeepLab \cite{chen2017deeplab} with and without post processing with CRF (denoted as DeepLab and DeepLab-CRF), CNN with encoder-decoder architecture \cite{nguyen2016detecting} on RGB and RGB-D images (denoted as ED-RGB and ED-RGBD), CNN with object detector (BB-CNN) and CRF (BB-CNN-CRF) \cite{nguyen2017object}. These baselines are already included in AffordanceNet, the current state-of-the-art architecture which we also compare our results to \cite{do2018affordancenet}. Additionally, we compare ourselves to the results obtained in an empirical investigation where authors replace the backbone of AffordanceNet and also test the importance of the multiple alignment feature \cite{minhlearning}. 

For the UMD dataset, we also report the results from the geometric features-based approach (HMD and SRF) \cite{myers2015affordance} and a deep learning-based approach that used both RGB and depth images as inputs (ED-RGBHHA) \cite{nguyen2016detecting}.

\section{Results}

We present the quantitative and qualitative results obtained in our experiments, and then draw conclusions on the relation between supervised object segmentation and supervised affordance segmentation. 

\subsection{Quantitative and qualitative evaluation}

\quad \textbf{Quantitative evaluation.} Our main results are given in Tables \ref{tb_result_iit} and \ref{tb_result_umd} (IIT-AFF and UMD datasets, respectively). We report the mean accuracy on each affordances classes, as well as a mean accuracy over all affordance classes, similarly to \cite{do2018affordancenet, minhlearning}. 

Our methods significantly surpass the state-of-the-art on both datasets. More specifically, we obtain a $5.2$\% and $11.5$\% increase over the second best performing method on the IIT-AFF and UMD datasets, respectively. We surpass the second best performing method on each affordance classes on both datasets, except the \texttt{grasp} category on the IIT-AFF dataset. The trained models that allow reproducing the results in this paper will be released upon acceptance.

\begin{figure}[!ht]
  \centering
{\includegraphics[scale=0.7]{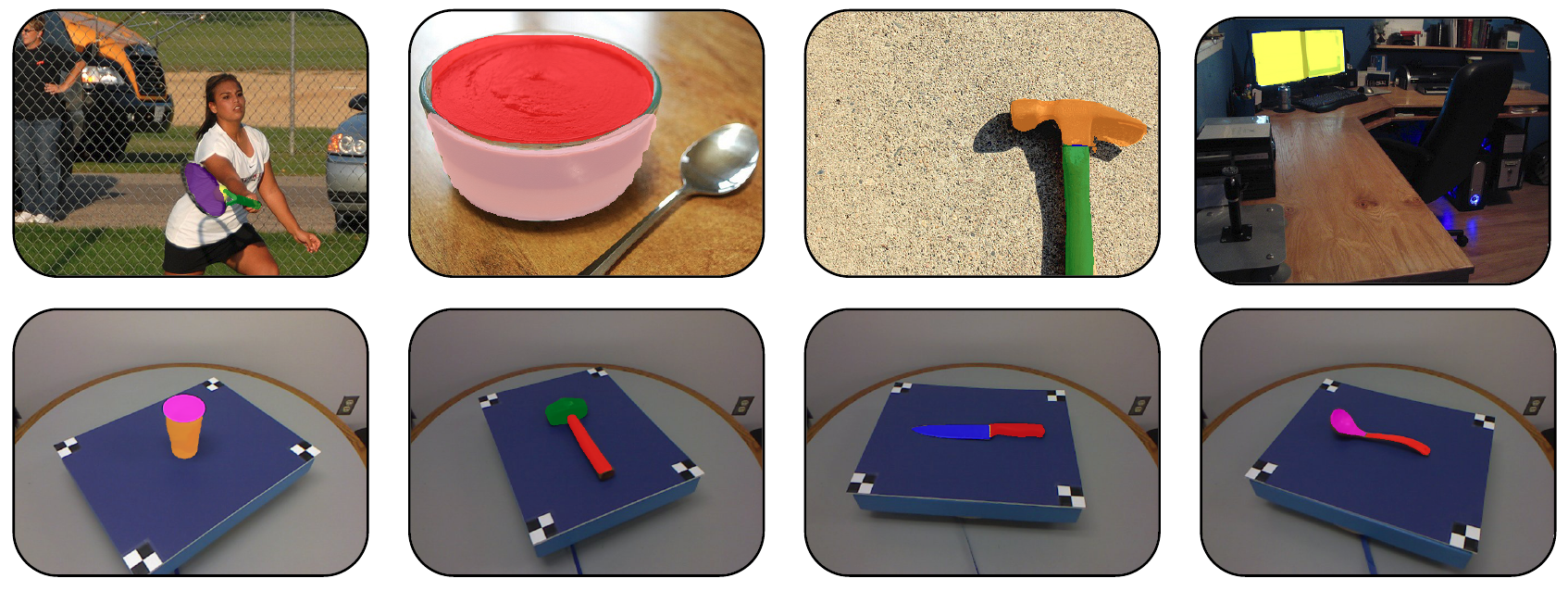}} \hfill
   \caption{Illustration predicted affordance segmentation masks using Mask R-CNN for the IIT-AFF \cite{nguyen2017object} (first row) and UMD \cite{myers2015affordance} (second row) datasets. \label{fig:res_illu2}}
\end{figure}

\quad \textbf{Qualitative evaluation.} Figures \ref{fig:res_illu} and \ref{fig:res_illu2} illustrate an uncurated selection of affordance segmentation masks predicted by Mask R-CNN. We can observe that the predicted masks are smooth, and correctly fit the ground-truth shape, without discontinuities or pixel artifacts. The method is able to predict masks of all shapes and sizes, with or without occlusion.

\begin{table}[!ht]
\centering
\caption{Performance on IIT-AFF Dataset ($F^{w}_{\beta} * 100$ metric, higher is better)}
\renewcommand\tabcolsep{1.9pt}
\label{tb_result_iit}
\hspace{2ex}

\begin{tabular}{@{}rccccccccccc@{}}
\toprule                                         &
{\shortstack{\small ED\\ \small RGB \\ \cite{nguyen2016detecting}} } &
{\shortstack{\small ED\\ \small RGBD \\ \cite{nguyen2016detecting}}} &
{\shortstack{\small DL\\ \small \\ \cite{chen2017deeplab}}} &
{\shortstack{\small DL\\ \small CRF \\ \cite{chen2017deeplab}}}  &
{\shortstack{\small BB\\CNN \small \\ \cite{nguyen2017object}}} &
{\shortstack{\small BB-CNN\\ \small CRF \\ \cite{nguyen2017object}}} &
{\shortstack{\small AN\\ \small VGG16 \\ \cite{do2018affordancenet}}} &
{\shortstack{\small AN\\ RN50 \\\cite{minhlearning}}} &
{\shortstack{\small AN\\ SE154 \\\cite{minhlearning}}} &
{\shortstack{\small Mask \\ R-CNN \\ (ours)}} &\\

\midrule
\texttt{contain}                                & 66.4   & 66.0   & 68.8     & 69.7 & 75.6     & 75.8   & 79.6 & 79.1  & 81.9 & \textbf{83.6} \\
\texttt{cut}                                    & 60.7   & 60.2   & 55.2     & 56.4 & 69.9     & 72.0   & 75.7 & 75.8 & 79.8 & \textbf{84.7}\\
\texttt{display}                                & 55.4   & 55.1   & 61.0     & 62.6 & 72.0     & 73.7   & 77.8 & 77.7 & 82.5 & \textbf{86.3} \\
\texttt{engine}                                 & 56.3   & 56.0   & 63.0     & 65.1 & 72.8     & 74.4   & 77.5 & 77.7 & 83.2 & \textbf{88.9}\\
\texttt{grasp}                                  & 59.0   & 58.6   & 54.3     & 56.2 & 63.7     & 64.3   & 68.5 & 69.0 & \textbf{77.1} & 71.1 \\
\texttt{hit}                                    & 60.8   & 60.5   & 58.4     & 60.2 & 66.6     & 67.1   & 70.8 & 70.8 & 79.2 & \textbf{92.3}\\
\texttt{pound}                                  & 54.3   & 54.0   & 54.3     & 55.5 & 64.1     & 64.9   & 69.6 & 69.7 & 78.2 & \textbf{81.8} \\
\texttt{support}                                & 55.4   & 55.0   & 54.3     & 55.6 & 65.0     & 66.1   & 69.8 & 70.0 & 78.9 & \textbf{86.7}\\
\texttt{w-grasp}                                & 50.7   & 50.4   & 56.0     & 57.5 & 67.3     & 68.4   & 71.0 & 71.1 & 80.6 & \textbf{83.7}\\
\midrule
\textbf{Average}                                & 57.6   & 57.3   & 58.4     & 59.9 & 68.6     & 69.6   & 73.4 & 73.4 & 80.2 & \textbf{84.4} \\
\bottomrule
\end{tabular}
\end{table}

\begin{table}[ht!]
\centering
\renewcommand\tabcolsep{2.7pt}
\caption{Performance on UMD Dataset ($F^{w}_{\beta} * 100$ metric, higher is better)}
\label{tb_result_umd}
\hspace{2ex}
\begin{tabular}{@{}rccccccccc@{}}
\toprule
& \shortstack{\small HMP\\ \small ~\cite{myers2015affordance}}
& \shortstack{\small SRF\\ \small ~\cite{myers2015affordance}}
& \shortstack{\small DL\\ \small ~\cite{chen2017deeplab}}
& \shortstack{\small ED-RGB\\ \small ~\cite{nguyen2016detecting}}
& \shortstack{\small ED-RGBD\\ \small ~\cite{nguyen2016detecting}}
& \shortstack{\small ED-RGB\\ \small HHA~\cite{nguyen2016detecting}}
& \shortstack{\small AN\\ \small \cite{do2018affordancenet}}
& \shortstack{\small Mask R-CNN\\ \small (ours)}
\\
\midrule
\texttt{grasp}                          & 36.7   & 31.4   & 62.0                             & 71.9                 & 71.4     & 67.3  & 73.1 & \textbf{78.9} \\
\texttt{w-grasp}                        & 37.3   & 28.5   & 73.0                             & 76.9                     & 76.7     & 65.2  & 81.4 & \textbf{93.6}\\
\texttt{cut}                            & 41.5   & 41.2   & 60.0                             & 73.7                         & 72.3     & 68.5  & 76.2 & \textbf{88.4} \\
\texttt{contain}                        & 81.0   & 63.5   & 90.0    & 81.7                         & 81.9     & 71.6  & 83.3  & \textbf{94.4}\\
\texttt{support}                        & 64.3   & 42.9   & 60.0                             & 78.0                         & 80.3     & 66.3  & 82.1 & \textbf{90.5} \\
\texttt{scoop}                          & 52.4   & 48.1       & 80.0     & 74.4                         & 75.7     & 63.5  & 79.3  & \textbf{85.4} \\
\texttt{pound}                          & 76.7   & 66.6   & 88.0    & 79.4                         & 80.6     & 70.1  & 83.6 & \textbf{92.7} \\ 
\midrule
\textbf{Average}                        & 55.7   & 46.0   & 73.3                             & 76.6                         & 77.0     & 67.5  & 79.9 & \textbf{89.1} \\
\bottomrule
\end{tabular}
\end{table}

\subsection{Conclusion: are standard Object Segmentation models sufficient for Learning Affordance Segmentation?}

Our results clearly show that Mask R-CNN is a well-suited algorithm for affordance segmentation mask prediction because it currently provides state-of-the-art performance on two commonly used datasets for this task. There is apparently no objective reason not to use Mask R-CNN, a method originally developed for object detection and segmentation, compared to methods tailored for this tasks such as \cite{do2018affordancenet}. Given the presented evidence, we conclude that for the moment, given the existing datasets and methods, the problem of supervised affordance segmentation is included in the problem of supervised object segmentation. In the following section, we discuss the potential reasons explaining this fact.

\section{Discussion}
\label{sec:discu}

We first discuss the importance of selecting strong baselines, and then analyse how affordance segmentation benchmarks could be improved.

\subsection{The importance of strong baselines}

The results we have obtained in this paper once again highlight the importance of selecting strong baselines when comparing methods on a benchmark. By improving the baselines, and comparing models in a fair way, the conclusions resulting from the experiments are often different. It is common to see papers that gives a sobering view on the state-of-the-art performance on a particular task. It has been the case with GANs \cite{lucic2017gans}, Transformers \cite{narang2021transformer}, Recommender Systems \cite{Dacrema_2019}, Disentanglement \cite{locatello2019challenging}, Neural Machine Translation \cite{denkowski2017stronger}, and the list goes on. These works have all been influential in their fields, providing an important perspective on the problem at hand, and fostering progress. When aggregated, these works also reinforce the idea that baseline selection is crucial for extracting meaningful results about the relative performance of algorithms on benchmarks. 

Our paper follows the same ideology: before tailoring new models to a particular task, we should first try to score as high as possible with standard methods. The first step is to identify re-usability of standard models. In the case of affordance segmentation, we identified that Mask R-CNN is directly applicable to the task. Even thought the task was not designed to be that similar to object segmentation, it turns out that object detection methods are sufficient to solve it.

\subsection{Limitations of supervised affordance segmentation benchmarks}

We argue that the main reason for the reported results is that the affordance segmentation benchmarks are too similar to object detection. In the case of the IIT-AFF and UMD datasets, the similarity with standard object segmentation datasets like COCO is too large. Since object detection in one the most studied field of Computer Vision, it seems unreasonable to think that it is possible to surpass those methods. This does not mean that supervised affordance segmentation is a useless task. The problem of affordances detection has many real-life application that surely go beyond simple object detection. It is more the way supervised affordance segmentation benchmarks are designed that do not allow to study the complex problem of affordances. 

Indeed, we possibly need different benchmarks that exploit characteristic properties of affordances. First, having only one possible class for each pixel in an image is not adapted to affordance segmentation, since one part of an object can have multiple affordances. It would be interesting to have a benchmark where a chair could be seen as an object on which both sitting and standing up is possible. In a related note, affordances are contextual: in different situations, objects do not afford the same thing. A pool can afford death for a baby, and also swimming for a person that can swim. Contextual affordance prediction and affordance relative to an individual given a context could be an interesting way to modify supervised affordance segmentation benchmarks. Recent effort have been pushing towards this interesting direction \cite{fang2018demo2vec, koppula2013learning, ardon2020affordances}. Notably \cite{chuang2018learning} proposes a benchmark with affordance segmentation that enable visual reasoning such as having the agent understand that there is a seat with someone currently sitting on it, meaning that the sitting affordance exists but is not currently available because of the context.

Finally, affordances are fundamentally tied to actions and behaviour. While this gets us further from supervised affordance segmentation, affordance detection benchmarks could consists in discovering the affordance of objects and scenes by having an artificial agent actually operate actions and learning about the affordances of the environment. There are several prior work that study this question \cite{zeng2018robotic, hamalainen2019affordance, liu2018physical, ardon2020affordances}. For instance, \cite{liu2018physical} attempts to predict object appearances and the effects of actions in order to understand the task. By doing so, they are able to predict the affordance of new objects through the decomposition of the object parts and use video frames to associate their effects when an action is applied. 

While progress is being made, as a community we need to construct proper simulation benchmarks where research teams can easily compare their approach to prior work, fostering significant progress as in object detection. For that we need to create benchmarks that focus on reproducibility and open-source access, while moving away for the supervised affordance segmentation task. Recent powerful simulators for indoor environments such as iGibson \cite{shen2020igibson} or AI2Thor \cite{ai2thor} allow to procedurally construct datasets and environments that can be shared among researchers. Using these simulators, many competitions are been proposed (e.g. iGibson 2021 challenge) where researchers can compare their approaches fairly. This could be an interesting experimental setup to create novel affordances benchmarks.

\section{Conclusion}

In this paper, we highlighted the sufficiency of standard object detection models, such as Mask R-CNN, for current supervised affordance segmentation benchmarks (ITT-AFF and UMD datasets). Our experiments show that Mask R-CNN is sufficient to significantly surpass the previous state-of-the-art performance, which was achieved by methods based on Mask R-CNN and specifically modified for supervised affordance segmentation. We infer that this surprising result is explained by the current limitations of the affordance segmentation benchmarks which are too similar to standard object segmentation datasets. We thus pointed several key aspects that benchmarks should have to accurately represent the affordance prediction task in its entirety: mainly by adding context and supporting actions and behaviour.

\bibliographystyle{plain}
\bibliography{bibli.bib}

\end{document}